\documentclass{IOS-Book-Article}
\usepackage[english]{babel}
\usepackage[utf8x]{inputenc}
\usepackage[T1]{fontenc}

\usepackage{mathptmx}
\usepackage{makeidx}  
\usepackage{lipsum}
\usepackage{multirow}
\usepackage[usenames, dvipsnames]{color}
\usepackage{graphicx}
\usepackage{wrapfig}
\usepackage{url}
\usepackage{hhline}
\usepackage{caption}
\usepackage{cleveref}
\usepackage{soul}
\usepackage{hyperref}
\usepackage{subcaption}
\usepackage{tikz} 
\usepackage{tikz-qtree-compat}  
\usepackage[colorinlistoftodos]{todonotes}
\usepackage{times}
\usepackage{hhline}

\def\hb{\hbox to 10.7 cm{}}

\begin{document}

\pagestyle{headings}

\begin{frontmatter}              


\title{Towards Egocentric Person Re-identification and Social Pattern Analysis}

\author[A,B]{\fnms{Estefania} \snm{Talavera}%
\thanks{Corresponding Author: Estefania Talavera, University of Groningen, Netherlands; E-mail:
e.talavera.martinez@rug.nl.}},
\author[B]{\fnms{Alexandre} \snm{Cola}},
\author[A]{\fnms{Nicolai} \snm{Petkov}}
and
\author[B]{\fnms{Petia} \snm{Radeva}}

\runningauthor{E. Talavera et al.}
\address[A]{Johann Bernoulli Institute of Mathematics and Computing Science, University of Groningen, The Netherlands}
\address[B]{Barcelona Perceptual Lab (BCNPCL). Dept. Matem\`atiques i Inform\`atica, University of Barcelona, Spain}

\begin{abstract}

Wearable cameras capture a first-person view of the daily activities of the camera wearer, offering a visual diary of the user behaviour. Detection of  appearance of people the camera user interacts with for social interactions analysis is of high interest. Generally speaking, social events, life-style and health are highly correlated, but there is a lack of tools to monitor and analyse them. We consider that egocentric vision provides a tool to obtain information and understand users social interactions. We propose a model that enables us to evaluate and visualize social traits obtained by analysing social interactions appearance within egocentric photostreams. Given sets of egocentric images, we detect the appearance of faces within the days of the camera wearer, and rely on clustering algorithms to group their feature descriptors in order to re-identify persons. Recurrence of detected faces within photostreams allows us to shape an idea of the social pattern of behaviour of the user. We validated our model over several weeks recorded by different camera wearers. Our findings indicate that social profiles are potentially useful for social behaviour interpretation.

\end{abstract}

\end{frontmatter}
\markboth{April 2018\hb}{April 2018\hb}


\section{Introduction}

Human social behaviour involves how people influence and interact with others, and how they are affected by others. This behaviour varies depending on the person, and is influenced by ethics, attitudes, or culture \cite{socialBackground}. Understanding behaviour of an individual is of high interest in social psychology. House et al. addressed the problem of how social relationships affect health \cite{socialrelationshipsandhealth} and demonstrated that social isolation leads to major risk factors for mortality. Moreover, Yang et al \cite{longevitysocial} observed that lack of social connections is associated with health risk in specific life stages, such as risk of inflammation in adolescence, or hypertension in old age. Also, as in Kawachi et al. \cite{socialties} it was highlighted, social ties have a beneficial effect in order to maintain psychological well-being.

\begin{figure}[ht!]
\includegraphics[width=1\textwidth,height=0.3in]{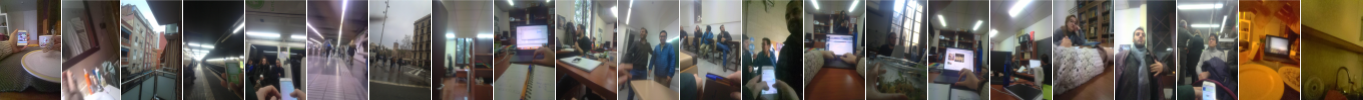}
\caption{Samples of egocentric photostream recorded by Narrative Clip camera along a day of the wearer of the camera.}
\label{Fig:pipeline}
\end{figure}

Considering the importance of the matter, automatic discovery and understanding of the social interactions are of high importance to the scientists, as they remove the need for manual labour. On the other hand, egocentric cameras are useful tools as they offer the opportunity to obtain images of daily activities of users from their own perspective. Therefore, providing a tool for automatic detection and characterization of social interactions through these recorded visual data can lead to personalized social pattern discoveries. 

In this work, we propose a method that enables us to answer questions such as \textit{Do I socialize along my days?} or \textit{With how many people do I interact daily?}. Given sets of egocentric images captured by several individuals, our proposed model employs a person re-identification model to achieve social pattern descriptions. First, a Haar-like feature-based cascade classifiers is applied \cite{Viola01rapidobject} to detect the appearing faces in the photo-streams. Detected faces in this step are converted to feature descriptors by applying the OpenFace tool \cite{amos2016openface}. Finally, we propose to define the person re-identification problem as a clustering problem. The clustering is applied over the pile of photostreams recorded by persons along the days to find the recurrent faces within photostreams. Shaping an idea about the social behaviour of the users becomes possible through referring to the time and day when the recurrences were appearing. 

\begin{figure}[ht!]
\includegraphics[width=1\textwidth,height=2in]{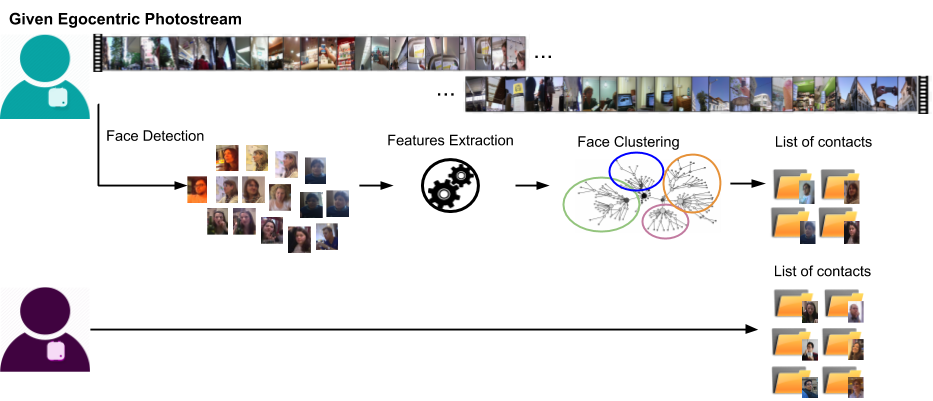}
\caption{Architecture of the proposed model. First, we apply the Viola\&Jones algorithm \cite{Viola01rapidobject} to detect appearing faces in the photostreams. Later, we apply the OpenFace tool to convert the faces to feature vectors. We propose to define the re-identification problem as a clustering problem with a later analysis of the grouped faces occurrence.}
\label{Fig:pipeline}
\end{figure}

The rest of the paper is organized as follows: in Section \ref{Section2:SOA}, we present an overview of the recent person re-identification approaches. In Section \ref{Section3:PersonReIdentification}, we define the Social Pattern Analysis approach, by performing face detection, faces clustering and evaluation of their occurrence. In Section \ref{Section4:Experiments}, we describe the experimental setup. Finally, in Section \ref{Section5:Discussion} we discuss our findings and draw conclusions and future research lines.


\section{State of the Art}
\label{Section2:SOA}

Social patterns analysis is commonly addressed as a re-identification problem. Re-identification usually considers information about detected faces in order to facilitate future detections. 

People re-identification is of high importance in the area of video surveillance, often identifying pedestrians recorded by different cameras \cite{Bak_2017_CVPR,Chen_2017_CVPR,Fan_2017_CVPR,Li_2017_CVPR,salienceLearning}. Security cameras generally capture the appearance of people, offering information such as clothes and pose. Challenges within this area of research include the changes in views, illumination and human body poses. 

Zhao et al. \cite{salienceLearning} proposed to incorporate salience maps and patch matching use of salience maps for unsupervised person re-identification. In \cite{Li_2017_CVPR}, Dangwei et al. introduced a deep neural network to learn and localize pedestrians, to later learn features from the body and parts of the body of the recorded people. Munawar et al. introduced a Convolutional Neural Network (CNN) to simultaneously register and represent faces \cite{Hayat_2017_CVPR}. Chen et al. addressed the problem of cross-camera variations and proposed a hashing based approach \cite{Chen_2017_CVPR}. They proposed to transform high-dimensional feature vector through binary coding scheme to compact binary codes that preserve identity. 

The methods introduced above evaluate images recorded by several surveillance cameras. These recorded images usually capture the body of the people and the local context. However, person re-identification from egocentric photostreams is different from person re-identification from datasets recorded by security cameras. On the other hand, egocentric images are commonly recorded by chest-worn cameras, they show social interactions from a closer perspective. Therefore, the face of the opposite person of the wearer commonly represents the main source of information about the person. 

Chenyou et al. targeted to stablish person-level correspondence across first- and third-person videos \cite{Fan_2017_CVPR}. To this end, they proposed a semi-Siamese CNN architecture. Aghaei et al. addressed social signal analysis from egocentric photostreams in \cite{Aghaei2017SocialPhoto-streams}. Here, the authors proposed to reach a characterization of the social pattern of a wearable photo-camera user through first, detection of social interactions, and later categorization of detected social interaction into either a formal or informal meeting. Through applying a face clustering algorithm, people forming the social environment of the user are localized throughout the social events of the user. This allows the social pattern characterization of the user through quantifying the density, diversity and social trends of the user.

This work goes one step forward and proposes to compare social patterns of different individuals by analysing their social behaviour from various aspects. After applying clustering of the detected faces, our proposed model quantifies the occurrence and duration of the interactions of the user with different individuals. Therefore, in this work we do not consider information about previously detected faces, but we rely on clustering techniques over the camera users data, in order to find the occurrence of the appearance of people around them. We consider that for our problem, this occurrence can be considered as re-identifying people along recorded days of the user.

\section{Social Patterns Characterization}
\label{Section3:PersonReIdentification}

People tend to interact with others along their day. By using wearable cameras, an egocentric perspective of those specific activities is captured. However, in order to address the analysis of social patterns of individuals, we first need to detect the appearance of people with whom the user interacts. Therefore, our approach towards social pattern analysis proposes to combine face detector, and face clustering algorithms.  

\subsection{Person Re-Identification}
\label{Section3:Subsection1:PersonReIdentification}

\textbf{Face detection} is performed by applying the well-known Viola-Jones algorithm \cite{Viola01rapidobject}. Our model translates the \textbf{re-identification} problem to a clustering problem. We rely on average linkage Agglomerative Hierarchical Clustering (AHC) to find the relation of the feature vectors  extracted from the detected faces. 

The consistency of the clusters is important for the later analysis of the occurrence of people within photostreams. We propose to estimate the clusters robustness by computing the Pearson's correlation coefficient among their samples. It is a measure of the linear relationship between two quantitative random variables, $x$ and $y$, that in our problem symbolize two feature vectors describing faces. Unlike covariance, Pearson's correlation $r_{xy}$ is independent of the measurement scale of the variables being determined as:

\begin{center}
$ r_{xy} = \frac{\sum x_{i}y_{i} - n\overline{x}\overline{y}}{(n-1)s_{x}s_{y}} = 
 \frac{n\sum x_{i} y_{i} - \sum x_{i} \sum y_{i}}{\sqrt{n\sum x_{i}^2 - (\sum x_{i})^2}\sqrt{n\sum y_{i}^2 - (\sum y_{i})^2}}. $
\end{center}
where $n$ is the sample size, $\overline{x}$ and $\overline{y}$ are the samples mean, and $xi$ and $yi$ are the single samples indexed with i.

The coefficient values have normalised values from -1 to 1, representing no similarity and identity, respectively. We discard an image within a cluster if the similarity coefficient does not reach a minimum value. The cluster consistency is checked when the mean of distance of the images feature descriptors is between 0.4 and 0.8. These empirical values are selected after running several experimental tests. A mean value higher than 0.8 is considered as robust. Inconsistent clusters were removed if the mean similarity value among their elements was less than 0.4. We defined a minimum similarity coefficient of 0.70. Clusters or images not following this consistency constraint are discarded and not later evaluated as social interactions.

When the clusters are found, and their consistency checked, we consider that people re-appear in the photostream when their faces belong to the same cluster. The temporal appearance of the people surrounding the camera wearer throughout photostreams describes the social pattern of behaviour of the wearable camera user. Therefore, after applying the clustering algorithm, the resulted groups can be analysed to find the occurrence and duration of relations.

\begin{figure}[ht!]
\includegraphics[width=1\textwidth]{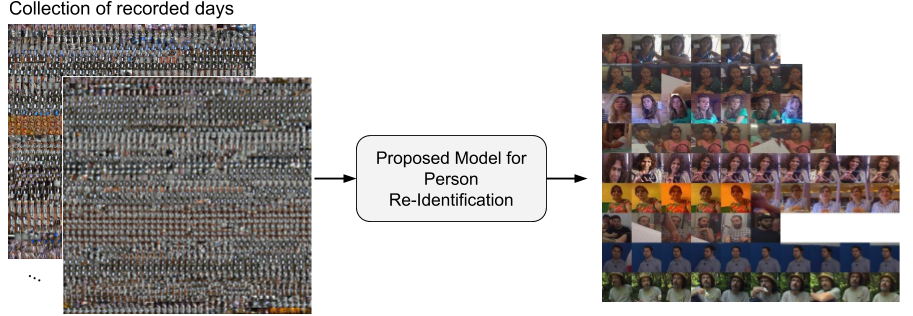}
\caption{Samples of the clusters obtained by applying our model for person re-identification on a set of recorded days by a camera wearer.}
\label{Fig:pipeline}
\end{figure}

We define several constraints when analysing the obtained clusterings due to the problem characteristics. On one hand, we seek sequential images with the appearance of a person. The wearable camera used for this experiment takes pictures every 20-30 second. We describe an event as a group of consecutive images of duration of a minimum of 3 minutes. Therefore, a cluster composed by images describing an event shorter than that time are not considered as an event, and consequently nor as social interaction of interest. On the other hand, the movement of the camera follows the movement of the user. This leads to challenging pose changes of the person, the camera wearer is interacting with, which makes faces detection and description challenging. Therefore, the appearance of faces can be either missed by the face detector algorithm, or not recognized by the neural network that extracts the face descriptors. Hence, we introduce a 15 minutes time-frame gap between temporal sequential images when calculating the duration of a social interaction.

\subsection{Social Profiles Comparison}
\label{Section3:Subsection2:Graphics}

A visual description of the extracted information is helpful when the aim is to interpret and compare social profiles of several users. This allows us to easily gain understanding of social behavioural differences among people. Since social patterns describe the camera wearer social interactions, a social pattern profile relates to  a single individual, describing a personalized social behaviour. Information such as \textit{time spent interacting with others}, or on the contrary, the \textit{time the user spent alone}, allow us to get insight of the social behaviour of a person. Although camera users do not wear the camera 24h per day, they wear it along  hours when most of the activities of their daily living occur. In this work, we propose the estimation of the following basic social characteristics through the analysis of the recorded hours of the camera users:

\renewcommand\labelitemi{$\cdot$}
\begin{itemize}
\item  \textit{Num p/day}: Average number of persons with whom the user interacts per days.

\item  \textit{Inter/day}: Average number of social interactions per day.

\item  \textit{T/P}: Average number of minutes spent with every person.

\item  \textit{T/Inter}: Average number of minutes spent per interaction.

\item  \textit{T/A}: Average number of minutes spent alone per day. 

\end{itemize}

\section{Experiments}
\label{Section4:Experiments}
\subsection{Dataset}

We collected data from 4 subjects who were asked to wear Narrative Clip camera (http://getnarrative.com/) fixed to their chest. This camera has a resolution of 2fpm and a normal lens. They captured information about their social daily routine,  taking pictures of the people with whom they interacted.  Since there is no training involved in this approach, the whole dataset is analysed by the proposed model. The dataset is organized as follows:
 
\renewcommand\labelitemi{$\cdot$}
\begin{itemize}
\item  \textit{User 1}: A set of 8766 images, composed by 1627, 1384, 395, 1490, 643, 1376, 1851 images per day, respectively.

\item  \textit{User 2}: A set of 10916 images, composed by 1395, 1587, 1926, 1615, 1643, 1371, 1379 images per day, respectively.

\item  \textit{User 3}: A set of 9053 images, composed by 729, 1601, 1055, 1753, 1302, 1434, 1179 images per day, respectively.

\item  \textit{User 4}: A set of 5343 images, composed by 780, 665, 747, 844, 809, 760, 792 images per day, respectively.
\end{itemize}

\subsection{Experimental setup}

We tested the face appearance detector over a test-set of 317 images, where 20 different person appear, and with a total of 377 faces. The best balance between performance and recognition accuracy rate was achieved configuring the \textit{scale factor} and \textit{minimum number of neighbours} parameters to 1.2 and 5, respectively. The algorithm achieved averages of 82.8\%, 56.6\% and 65\% rates for Precision, Recall, and F-score, respectively. Later, we use the OpenFace tool \cite{amos2016openface}, a trained deep neural network that extracts 128-D feature vectors from the detected faces.

After detecting the appearance of faces in the egocentric photostreams and extracting their descriptors, the clustering method is applied to find their relation and temporal occurrence. Our proposed clustering algorithm is evaluated over an extended version of the test-set, composed by 4280 egocentric images. We compared its performance to the state-of-the-art MeanShift \cite{meanshift} and Spectral Clustering \cite{spectralclustering} techniques as well as with other configuration of agglomerative clusterings, with different dissimilarity metrics. The robustness for the obtained clusters is considered for all the methods evaluated. We evaluate the obtained results with Precision, Recall, and F-Measure metrics, see Table \ref{resultsClustering}.

\begin{table}[ht!]
\centering
\caption{Average Precision, Recall, and F-Measure result for each of the tested methods on the extended test-set composed by egocentric images.}
\label{resultsClustering}
\resizebox{\textwidth}{!}{\begin{tabular}{lcll|cll|cll}
 & \multicolumn{3}{c|}{Proposed clustering model} & \multicolumn{3}{c|}{Mean Shift} & \multicolumn{3}{c}{Spectral Clustering} \\ \cline{2-10} 
 & \multicolumn{1}{l}{Precision} & Recall & F-Measure & \multicolumn{1}{l}{Precision} & Recall & F-Measure & \multicolumn{1}{l}{Precision} & Recall & F-Measure \\ \hline
\multicolumn{1}{l|}{Extended Test-set} & 81,66 & \multicolumn{1}{c}{33,48} & \multicolumn{1}{c|}{44,14} & 37,69 & \multicolumn{1}{c}{15,47} & \multicolumn{1}{c|}{21,70} & 93,50 & \multicolumn{1}{c}{29,74} & \multicolumn{1}{c}{42,47} \\ \hline \hline
\end{tabular}}
\end{table}

We expected that the clusters relate to different people that the wearer interacted with. Finally, information about the individual interactions is derived through the  evaluation of the faces occurrence along the day.

\subsection{Results}

\begin{figure}[ht!]
\centering

\includegraphics[width=.5\textwidth]{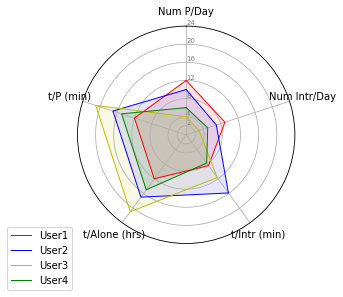} 

\medskip
\includegraphics[width=.45\textwidth]{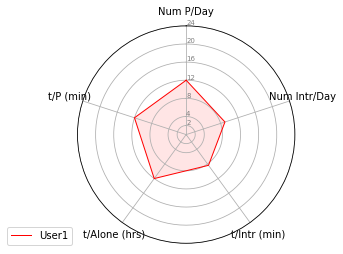}
\includegraphics[width=.45\textwidth]{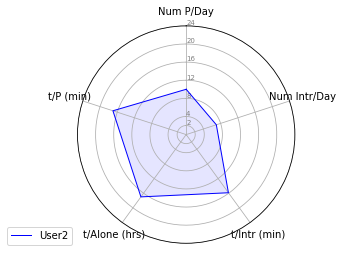}\quad
\medskip
\includegraphics[width=.45\textwidth]{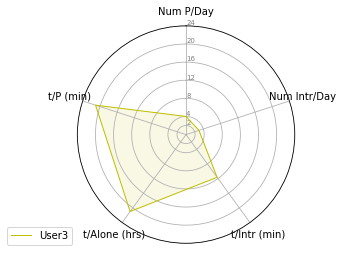}
\includegraphics[width=.45\textwidth]{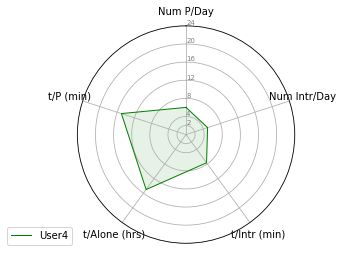}

\caption{Obtained social profiles as a result of applying our method to egocentric photostreams recorded by different users. For a better interpretation, we present a social profile per user, and a common one where the social profiles are overlapping. The last one offers a clearer view of differences among individuals.}
\label{pics:socialProfile}
\end{figure}

The obtained results are reported both, numerically in Table \ref{Table:socialtraits} and visually, through social profiles in Fig. \ref{pics:socialProfile}.

\begin{table}[ht!]
\centering
\caption{This table shows the social behavioural traits obtained from the detected social interactions for the different camera wearer.}
\label{Table:socialtraits}
\begin{tabular}{llllll}
 & \multicolumn{5}{c}{Social Behavioural Traits} \\ \cline{2-6} 
 & Num p/day & Avg int/day & Avg t/int (min)& Avg t/p (min)& Avg t/alone (h)\\ \hline \hline
\multicolumn{1}{l|}{User 1} & 9 & 12 & 12 & 12 & 8h 23m \\
\multicolumn{1}{l|}{User 2} & 7 & 10 & 17 & 17 & 15h 52min \\
\multicolumn{1}{l|}{User 3} & 3 & 4 & 21  & 21 & 11h 39min \\
\multicolumn{1}{l|}{User 4} & 5 & 6 & 15 & 15 & 7h 42min \\ \hline \hline
\end{tabular}
\end{table}

In Fig. \ref{pics:socialProfile}, we can observe how social patterns differ among individuals. For instance, User 2 interacts with a lower number of people per day than User1, but the duration of those interactions is higher. On the other hand, User 3 is the individual that spends a higher average time per person, even though he interacts with the lower number of people per day. We can infer from that that his social interactions are with on a small group of people. 

Qualitative results suggest that the presented automatic model for social patterns analysis rapidly obtains social behaviour understanding of the individuals. Therefore, further interpretations of social profiles are on social specialists hands. Moreover, on the weeks of egocentric sets, a total of 182 clusters were obtained: 61, 63, 16, and 42, for User1, User2, User3 and User4, respectively.


\section{Conclusions}
\label{Section5:Discussion}

In this work, we proposed a model to automatically analyse social behaviour of wearable cameras users, through quantifying and visualizing the occurrence of their social interactions. This is of high interest for social psychology, due to the fact that it removes the need for manual labour when analyzing social behaviour of people. To this end, we rely on clustering algorithms to find the relation of detected appearance of people throughout the egocentric photostreams. 
We propose to obtain five social characteristics about the detected social interactions. These social traits are used to create a social profile through their visualization. Social profiles allow scientists to obtain information and compare different social behaviour of individuals, for its later study and understanding.
Currently, this model and the obtained results are under discussion with social psychologists. Their comments about relevant traits to be obtained from the recorded photostreams will infer a more robust social profile.
Future work will address the problem of describing the kind of relation the camera wearers share with the people they interact with. An application will be to use the recognized contacts of the user for cognitive training of patients suffering from Alzheimer disease.

\newpage

\section*{Acknowledgments}
This work was partially founded by Ministerio de Ciencia e Innovaci\'on of the Gobierno de Espa\~na, through the research project TIN2015-66951-C2. SGR 1219, CERCA, \textit{ICREA Academia 2014} and Grant 2014-1510 (Marat\'{o} TV3). The funders had no role in the study design, data collection, analysis, and preparation of the manuscript.

\bibliographystyle{plain}

\end{document}